# Ensemble Hyperspectral Band Selection for Detecting Nitrogen Status in Grape Leaves

Ryan Omidi
*Department of Electrical and Computer Engineering*
*California State Polytechnic University, Pomona*
Pomona, USA
raomidi@cpp.edu

Ali Moghimi
*Department of Biological and Agricultural Engineering*
*University of California, Davis*
Davis, USA
amoghimi@ucdavis.edu

Alireza Pourreza
*Department of Biological and Agricultural Engineering*
*University of California, Davis*
Davis, USA
apourreza@ucdavis.edu

Mohamed El-Hadedy
*Department of Electrical and Computer Engineering*
*California State Polytechnic University, Pomona*
Pomona, USA
mealy@cpp.edu

Anas Salah Eddin
*Department of Electrical and Computer Engineering*
*California State Polytechnic University, Pomona*
Pomona, USA
asalaheddin@cpp.edu

*Abstract*—The large data size and dimensionality of hyperspectral data demands complex processing and data analysis. Multispectral data do not suffer the same limitations, but are normally restricted to blue, green, red, red edge, and near infrared bands. This study aimed to identify the optimal set of spectral bands for nitrogen detection in grape leaves using ensemble feature selection on hyperspectral data from over 3,000 leaves from 150 'Flame Seedless' table grapevines. Six machine learning base rankers were included in the ensemble: random forest, LASSO, SelectKBest, ReliefF, SVM-RFE, and chaotic crow search algorithm (CCSA). The pipeline identified less than 0.45% of the bands as most informative about grape nitrogen status. The selected violet, yellow-orange, and shortwave infrared bands lie outside of the typical blue, green, red, red edge, and near infrared bands of commercial multispectral cameras, so the potential improvement in remote sensing of nitrogen in grapevines brought forth by a customized multispectral sensor centered at the selected bands is promising and worth further investigation. The proposed pipeline may also be used for application-specific multispectral sensor design in domains other than agriculture.

## I. Introduction

Hyperspectral data, with their fine spectral resolution, are used extensively in remote sensing applications such as vegetation species classification and estimation of leaf area index, carbon and nitrogen content, and plant stress [1]. While multispectral sensors normally have several broad (e.g. 60 nm) wavebands, hyperspectral sensors can have hundreds or thousands of wavebands, each with narrow bandwidth (e.g. 1 nm, 5 nm, or 10 nm) [2]. Compared to broad spectral bands, hyperspectral bands have been shown to account for 25% more variability in crop discrimination [3]. However, hyperspectral data have limitations relative to multispectral data including handling large data volumes and massive data processing [2], along with data redundancy [4]. Hyperspectral sensors can generate terabytes to exabytes of data [5], which may make storage or processing intractable. The high dimensionality, data size, and redundancy can make it difficult to identify which wavelengths are relevant and useful for analysis [5]. Amid the hundreds of wavelengths measured by a hyperspectral sensor, only a small set of wavelengths may be related to the target traits, while the remaining wavelengths are often irrelevant or redundant, making interpretation difficult and increasing the risk of overfitting [6]. In fact, the large number of measured wavebands can harm the classification performance due to the curse of dimensionality or Hughes effect [7], which is the loss of classifiability observed when the dimensionality of the data increases while the number of training samples remains fixed [8]. Since the size of the training data can be restricted by time and budgetary constraints, the data dimensionality can be reduced using feature selection methods.

Pal and Foody showed that the SVM-RFE, CFS, mRMR, and random forest feature selection techniques improved the accuracy of an SVM classifier when detecting different land-cover types in hyperspectral datasets [9]. Interestingly, Hennessy, Clarke, and Lewis found considerable variability in the optimal wavebands selected by different feature selection techniques for hyperspectral plant classification and therefore recommended the use of multiple classification models or an ensemble of feature selectors for evaluation by a single classifier [10]. Ensemble feature selection combines the results from several base feature selection methods, or rankers, into an aggregated final ranking [11]. This integration of results from a diverse set of base rankers has been shown to outperform the popular SVM-RFE base ranker in wide gene-expression datasets [12].

Hyperspectral data often exhibits high correlations between adjacent, redundant bands [5]. Some feature selection methods may struggle with highly correlated features due to correlation bias, which dilutes the importance of features in large correlated groups, but can be improved by using a single representative feature from each group of correlated features [13, 14]. Though multispectral data are not affected by the same limitations as hyperspectral data, commercial UAV-mounted multispectral cameras typically perform measurements in the blue, green, red, red edge, and near infrared bands, mirroring common bands used by modern satellite remote sensing platforms like RapidEye, WorldView-2, and others [15]. These bands may or may not be optimal for a given application, e.g. performing nitrogen detection in grape leaves.

The large amount of redundant bands measured by hyperspectral sensors mean that complex processing and data mining become necessary in order to make practical use of the data collected. The blue, green, red, red edge, and near infrared bands measured by typical commercial multispectral sensors can be limiting, since they may not be optimal for every application. Feature selection techniques for hyperspectral data typically select a subset of the original feature wavelengths, which can be quite narrow in bandwidth (e.g. < 10 nm full width at half maximum), while typical drone-mounted multispectral sensors have bandwidths ranging 10 nm - 40 nm. Combining feature selection and band clustering techniques on hyperspectral data may reveal optimal spectral bands beyond the common blue, green, red, red edge, and near infrared bands for performing nitrogen detection in grape leaves. A feature selection pipeline which identifies an optimal set of bands suitable for use in drone-mounted multispectral sensors would aid in the design of customized application-specific multispectral sensors which do not require a data mining step before practical usage.

## II. METHODS

### A. Data Collection

The data were collected from 'Flame Seedless' grapevines in a commercial table grape vineyard in Kingsburg, California. Vines in the study area either remained non-fertilized, or received different amounts of nitrogen in order to induce a range of tissue nitrogen content among the vines. Approximately 20 leaves were collected from each of the 150 vines and were then subjected to hyperspectral measurements. Hyperspectral measurements from each leaf sample were collected with two spectrometers: Flame-S (186 - 1031 nm with 0.4 nm resolution, Ocean Optics) and Flame-NIR (936 - 1660 nm with 6 nm resolution). A Spectralon Diffuse Reflectance Standard (Labsphere) was used to calibrate the raw data read by the spectrometers and to produce reflectance values. After hyperspectral measurements were made, the leaves were dried in a forced air oven, ground into a fine powder, and then submitted to a commercial laboratory (Dellavalle Laboratory, Inc., Fresno, CA) for determination of total nitrogen. This generated, for each of the 150 vines, one true nitrogen label expressed as a percentage of total dry mass.

### B. Ensemble Design

For filter rankers, SelectKBest [16] was chosen because it is a standard technique used to identify features with large variance [17]. It is part of the popular scikit-learn machine learning package for Python and has a small computation time [18]. ReliefF [19] was chosen because it is good at detecting conditional dependencies and is noise tolerant [20, 21, 22]. For wrapper rankers, SVM-RFE [23] was chosen because it is faster and less prone to overfitting than many other wrapper methods and has been successfully applied in domains like gene selection involving many features and few samples [13]. The Chaotic Crow Search Algorithm (CCSA), based on the Crow Search Algorithm [24], was chosen because it is a relatively new and popular method which has been shown to outperform other popular meta-heuristic algorithms in feature selection [25]; the author's MATLAB open source code [26] was ported to Python and modified to produce feature rankings for integration into this study's pipeline. Each feature's CCSA ranking was determined by its lifetime in top scoring feature sets, e.g. if the top scoring feature set of every round included "1619 nm", then "1619 nm" would have a high ranking. For embedded rankers, LASSO [27] was chosen because it tends to produce sparse solutions and is useful when the number of features is large relative to the number of samples. Random Forest [28] was chosen because it is noise tolerant, does not overfit, incorporates feature dependencies, and does not require fine tuning [29, 30].

### C. Data Preparation

Hyperspectral reflectance and nitrogen data were loaded into a pandas DataFrame in Python. The nitrogen label for each vine was replicated for each leaf from the same vine. For the Flame-S sensor, wavelengths below 400 nm and above 900 nm were observed to be noisy and were removed from the DataFrame. The reflectance data from both Flame-S and Flame-NIR sensors were merged into one dataset. Outlier samples containing one or more reflectance values over three standard deviations from the mean reflectance for the corresponding wavelengths were removed from the dataset.

TABLE I. NITROGEN THRESHOLDS

| Region (class) | Nitrogen % | Samples |
| --- | --- | --- |
| Low (0) | N <= 2.55 | 346 |
| Inner Low (0) | 2.55 < N < 2.66 | 127 |
| Inner High (1) | 3.35 < N < 3.4 | 101 |
| High (1) | N >= 3.4 | 374 |

Since hyperspectral samples were collected per leaf and ground truth labels were collected per vine using the industry standard method, this study was modelled as a classification of extreme cases as a proof of concept rather than as a regression problem, i.e. the classifier determined whether each leaf belonged to a vine which was deficient or sufficient in nitrogen rather than estimating the average nitrogen level of the whole vine based on individual leaves. Four nitrogen thresholds (Table I) were chosen in order to maintain balance in sample numbers and separation between class 0 (low nitrogen) and class 1 (high nitrogen). This segmented the dataset into two extreme nitrogen regions, two inner nitrogen regions, and a middle nitrogen region. The extreme samples were used as a training set, the inner samples were used as a test set, and the middle samples were not used because their class was ambiguous. The hyperspectral data from the low and high samples were denoted as $X_{extreme}$ and the corresponding nitrogen labels from the low and high samples were represented as class 0 or 1, respectively, and denoted as $y_{extreme}$. The hyperspectral data from the inner low and inner high samples were denoted as $X_{inner}$ and the corresponding nitrogen labels from the inner low and inner high samples were represented as class 0 or 1, respectively, and denoted as $y_{inner}$.

*D. Band Pair Correlation*

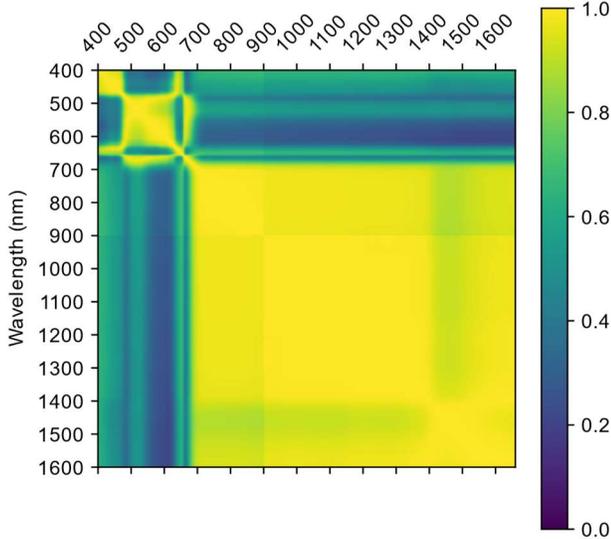

Fig. 1. Pairwise correlation between bands as a symmetric colormap image. Flame-S wavelengths spanned from 400 to 900 nm and Flame-NIR wavelengths spanned from 936 to 1660 nm.

Hyperspectral wavelengths are generally close together and can display high correlation with adjacent wavelengths. In this study, bands centered between ~700 and ~900 nm and ~936 and ~1350 nm were highly correlated (Fig. 1), which indicated that some of these wavelengths were redundant and likely to be averaged together during correlation windowing described below. The dark blue stripe between ~550 and ~620 nm and the light green stripe between ~1400 and ~1500 nm indicated these bands were less correlated with neighboring wavelengths. In order to mitigate correlation bias and help the rankers identify the most informative windows of wavelengths with which to start the selection process, correlated adjacent wavelengths were grouped and averaged together using a tunable correlation threshold to produce a single representative feature from each highly correlated group. The correlation matrix of the training set $X_{extreme}$ was used to window both itself and the test set $X_{inner}$ to avoid leaking knowledge about the test set and to ensure the features matched. Averaged windows of wavelengths were renamed with the midpoint wavelength of the window. The output of the correlation windowing step was $X_{extreme\ (windowed)}$, $X_{inner\ (windowed)}$, and a dictionary object containing the first, last, and midpoint wavelengths from each window. A very high correlation threshold of 0.99 was used in this study, which still managed to drop the number of features from 1,339 initial wavelengths to 51 correlated window features. The resulting datasets, $X_{extreme\ (windowed)}$ and $X_{inner\ (windowed)}$, were normalized to have a mean of zero and standard deviation of one for each feature. The original mean and standard deviation of each feature from the training set $X_{extreme\ (windowed)}$ was used to normalize the features in both the training and test sets to avoid leaking knowledge about the test set. The outputs of normalization were denoted as $X_{extreme\ (windowed,\ normalized)}$ and $X_{inner\ (windowed,\ normalized)}$. Averaging highly correlated adjacent wavelengths initially reduced the number of features to select from and helped the classifier by mitigating collinearity between features. It helped to heuristically identify the most informative windows of wavelengths in the spectrum to focus further searching for an optimal feature set.

*E. Ensemble Feature Selection Pipeline*

Each of the six base rankers produced rankings for the features in $X_{extreme\ (windowed,\ normalized)}$. These base rankings were used in a recursive ranker elimination process described in [6] in which the best performances of subsets produced by an ensemble of 6, 5, 4, 3, 2, and 1 ranker were compared to determine a winning feature subset. The decision criterion in [6] was TOPSIS, which was replaced in this study with a fitness function composed of mean cross validation F1 score and subset smallness. The output of this stage was the set of top correlated window features.

Since the goal of the study was to identify an optimal set of bands for leaf nitrogen classification suitable for multispectral cameras, a center wavelength for each 10, 20, or 40 nm wide band was desired. Some features within the correlated windows were found to be more informative than others. Since the highly correlated features comprising each correlated window were previously averaged together, this stage required the initial non-windowed dataset with original wavelength features and a correlation window dictionary to locate the first and last wavelengths in each window. For each of the top correlated windows, the rankings of the wavelengths within the window were found. Recursive ranker elimination was performed again to produce the best set of wavelengths within the window. The center of a new 10, 20, or 40 nm wide band was first placed at the top ranked wavelength in this set. The center was averaged towards subsequent wavelengths in the set using a clustering algorithm to contain as many top ranked wavelengths as possible in the clustered band. Re-clustering was performed on overlapping bands depending on their bandwidths and degree of overlap; a 40 nm band overlapping any other band re-clustered to a new 40 nm band centered between them, a 10 nm band overlapping a 20 nm band re-clustered to a new 20 nm band centered between them, two 20 nm bands overlapping were rounded to either a new 20 nm or new 40 nm band depending on the distance between their centers, and two 10 nm bands overlapping were rounded to either a new 10 nm or new 20 nm band depending on the distance between their centers. The centers and lower and upper bounds of each clustered band were stored in a clustered window dictionary object and passed to the data preparation function to re-window the wavelengths in the dataset to match the clustered window bands.

At this stage of the pipeline, the clustered window features were just constructed and had not been evaluated on the training data or compared with each other, so a final stage of ensemble feature selection was needed to check for the best set of clustered window features. Base rankings were found for the clustered window features and recursive ranker elimination was run once again to find the best set of clustered window features or bands.

A QDA classifier was trained on $X_{extreme\ (clustered,\ normalized)}$ and $y_{extreme}$ using only the top clustered window bands, which produced a class-weighted F1 classification score with a mean and standard deviation over the 10 cross validation folds. The trained classifier made predictions of nitrogen level on the unseen moderate nitrogen samples, $X_{inner\ (clustered,\ normalized)}$, using only the top clustered window bands as input. A class weighted F1 score was calculated based on the test labels, $y_{inner}$. The full pipeline is visualized in Fig. 2. F1 scores on the training and test sets were also calculated using the full set of correlated windows, the set selected by the full ensemble, the sets selected by the individual base rankers, and the set selected by the winning ensemble from one stage of recursive ranker elimination for comparison.

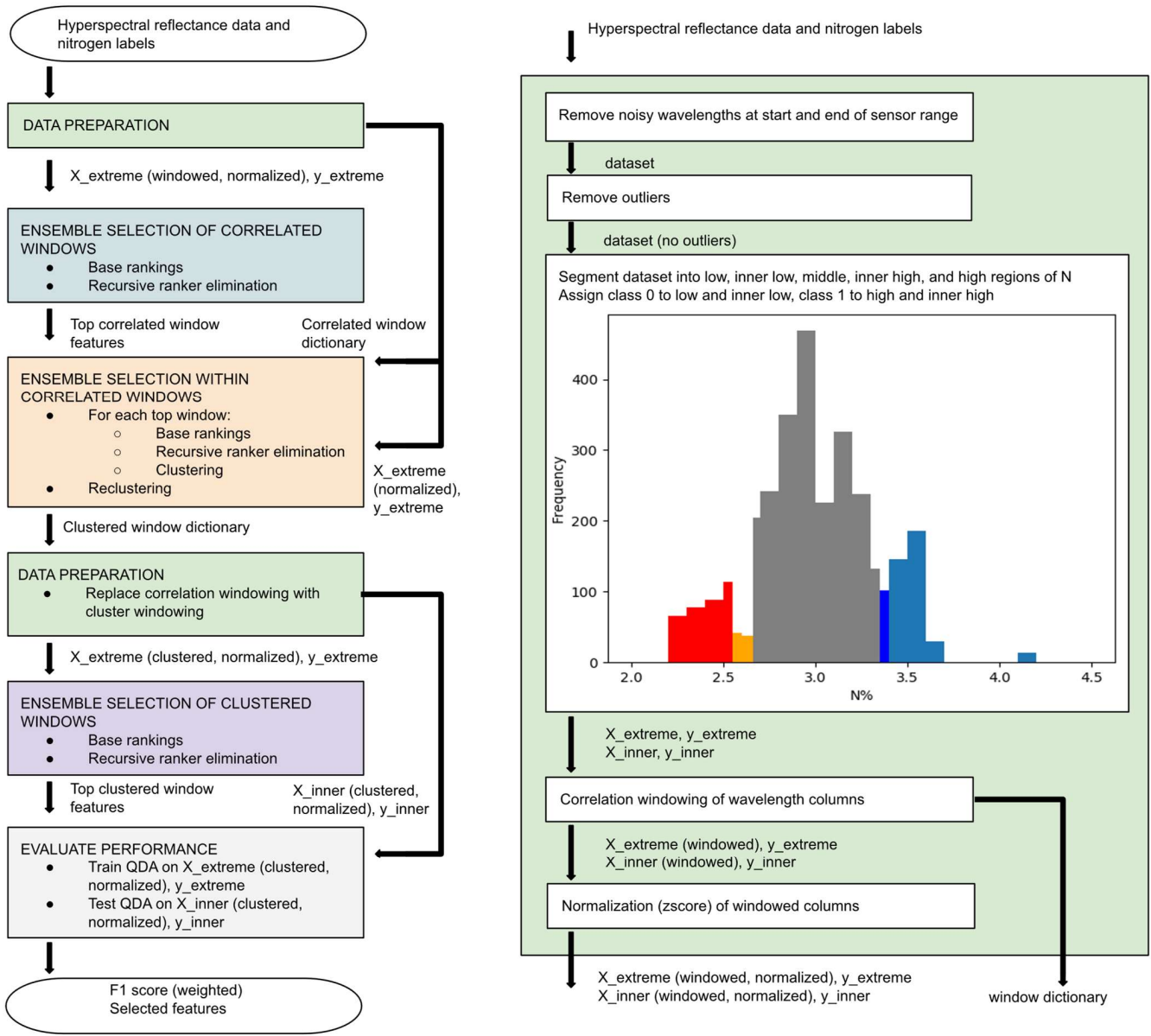

Fig. 2. Ensemble feature selection pipeline flowchart (left). Data preparation flowchart (right).

## III. RESULTS

### A. Correlated Window Ranking

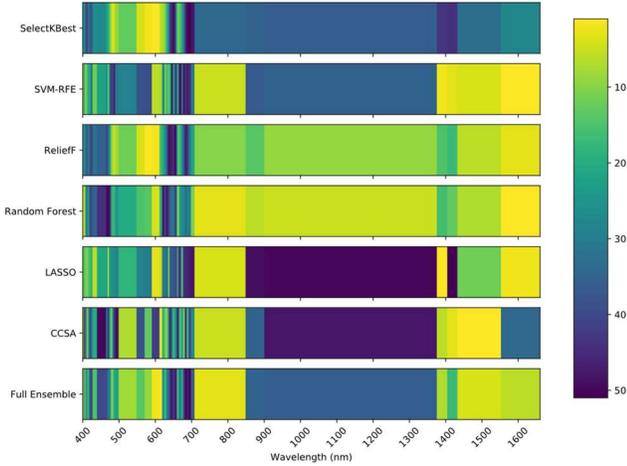

Fig. 3. Heatmaps of spectral feature rankings as determined by each individual base ranker and the average of all their feature rankings (Full Ensemble). Brighter colors indicate a higher ranking. Each single-colored block represents a range of highly correlated ($\rho > 0.99$) wavelengths averaged together during correlation windowing into one single feature.

A high correlation threshold was selected to mitigate the chance of grouping and averaging two or more highly informative features into the same correlation window and treating them as a single averaged feature, effectively losing valuable information. The potential risk of increasing feature redundancy by choosing a high correlation threshold was addressed with the subsequent stages of feature selection in the pipeline. Filter methods SelectKBest and ReliefF determined the most informative features for grape nitrogen classification were centered at ~580 nm and ~601 nm (yellow-orange light), while LASSO also identified ~601 nm as highly informative (Fig. 3). However, CCSA and Random Forest did not determine the ~580 nm or ~601 nm features to be highly informative, which underlined the benefit of using an ensemble of ranking methods in order to detect informative features which some individual rankers may miss. Similarly, though the filter methods did not determine the feature centered at ~1392 nm to be highly informative, wrapper methods SVM-RFE and CCSA both identified this feature as highly informative. The full ensemble identified the features centered at ~615, ~601, ~778, ~1494, ~401, ~1608, and ~1392 nm, respectively, as most informative.

### B. Recursive Ranker Elimination

The first ranker eliminated was SelectKBest, followed by SVM-RFE, CCSA, ReliefF, and finally Random Forest (Table II). The subset selected by the last remaining ranker, LASSO, contained 8 features and produced a mean cross validation F1 score of about 0.88 on the extreme nitrogen training set, resulting in a fitness score of about 1.72, which was outperformed by the subset of 7 features selected by the ensemble of 3 rankers: Random Forest, LASSO, and ReliefF. The winning ensemble from the entire six rounds of recursive ranker elimination was the ensemble of 3 rankers: {'Random Forest', 'LASSO', 'ReliefF'}. The feature subset it selected contained only 7 of the 51 total correlated window features, centered at 1608, 601, 778, 1494, 1392, 401, and 615 nm, respectively. With these 7 features, a QDA classifier was able to obtain a mean cross validation F1 score of about 0.87 on the extreme nitrogen training set. The best base ranker, i.e. the last one standing from recursive ranker elimination, was LASSO. LASSO's selected subsets of sizes 2, 3, 4, 6, 8, 9, 10, and 11 were observed to outperform those selected by the full ensemble and by the winning ensemble of 3 rankers, but its optimal subset of 8 features was defeated by the winning ensemble's optimal subset of 7 features based on fitness score.

TABLE II. RECURSIVE RANKER ELIMINATION RESULTS

| Ensemble size (rankers) | Rankers remaining | Subset size (features) | F1 mean | F1 std | Fitness score |
|---|---|---|---|---|---|
| 6 | {'SVM-RFE', 'SelectKBest', 'CCSA', 'LASSO', 'Random Forest', 'ReliefF'} | 7 | 0.86 | 0.03 | 1.73 |
| 5 | {'SVM-RFE', 'CCSA', 'LASSO', 'Random Forest', 'ReliefF'} | 7 | 0.86 | 0.01 | 1.73 |
| 4 | {'Random Forest', 'LASSO', 'ReliefF', 'CCSA'} | 7 | 0.86 | 0.04 | 1.73 |
| 3 | {'Random Forest', 'LASSO', 'ReliefF'} | 7 | 0.87 | 0.03 | 1.73 |
| 2 | {'Random Forest', 'LASSO'} | 8 | 0.86 | 0.05 | 1.71 |
| 1 | {'LASSO'} | 8 | 0.88 | 0.03 | 1.72 |

### C. Within-Window Selection and Clustering

TABLE III. CLUSTER WINDOWING RESULTS

| Correlation-windowed features | | | Cluster-windowed features | | |
|---|---|---|---|---|---|
| Center (nm) | Range (nm) | Bandwidth (nm) | Center (nm) | Range (nm) | Bandwidth (nm) |
| 1608 | 1557 - 1659 | 102 | 1619 | 1596 - 1642 | 46 |
| 601 | 590 - 611 | 21 | 596 | 586 - 606 | 20 |
| 778 | 707 - 848 | 141 | 825 | 805 - 845 | 40 |
| 1494 | 1437 - 1551 | 114 | 1528 | 1505 - 1551 | 46 |
| 1392 | 1380 - 1403 | 23 | 1386 | 1374 - 1397 | 23 |
| 401 | 400 - 402 | 2 | 401 | 400 - 406 | 6 |
| 615 | 611 - 618 | 7 | 611 | 606 - 616 | 10 |

The next stage of the pipeline looked within the top correlated windows selected by the last stage, centered at 1608, 601, 778, 1494, 1392, 401, and 615 nm. The range of the original, highly correlated, wavelengths which were averaged together to create each of these features were similarly run through recursive ranker elimination. The optimal subset of wavelengths from within each window was used in a clustering algorithm to choose the best place for the center of a ~10 nm, ~20 nm, or ~40 nm wide clustered window. The smooth color gradient found in part of the LASSO heatmap (not shown) indicated that some of the highly correlated ($\rho > 0.99$) features within the 1608 nm centered correlated window were difficult for LASSO to rank so their original orders were left intact. The 5 optimal features selected by the winning ensemble of 2 rankers, {'SVM-RFE', 'LASSO'}, were clustered to center a ~40 nm wide spectral band. This was repeated for the remaining six correlated windows from the last stage's winning subset. Correlated window bandwidths varied quite a bit (Table III) since they were dynamically chosen according to the correlation matrix of the hyperspectral dataset, whereas clustered windows had more consistent bandwidths close to the typical bandwidths of 10, 20, or 40 nm found in commercial multispectral cameras. The variations from exactly 10, 20, or 40 nm bandwidth are accounted for by the Flame-NIR sensor's spectral resolution of ~6 nm and the lower limit of the Flame-S sensor having been clipped at 400 nm. A second round of clustering was performed on the clustered windows themselves to prevent overlapping clustered windows originating from different correlated windows.

### D. Clustered Window Selection

The final stage of feature selection was performed on the clustered windows centered at 401, 596, 611, 825, 1386, 1528, and 1619 nm from the previous stage, in order to compare the newly constructed features for the first time and to determine the optimal subset of these features for grape nitrogen estimation. The feature subset selected by the winning ensemble for this stage contained the six bands centered at 596, 611, 401, 1619, 825, and 1386 nm.

### E. Nitrogen Estimation

Without any feature selection, using the full feature set of 51 correlated window features allowed the QDA classifier to obtain a mean F1 score of about 0.93 (Table IV) on the extreme nitrogen samples in the training set, and about 0.69 on the moderate nitrogen samples in the test set. The best of the base rankers were SVM-RFE, LASSO, and CCSA, each obtaining mean training F1 scores of at least 0.85 and test F1 scores of at least 0.67 with only 10 or fewer correlated window features. The full ensemble of six rankers obtained a similar mean F1 score with only 7 features, while the one-stage recursive ranker elimination obtained a slightly better mean F1 score and smaller standard deviation than the full ensemble did. The full pipeline consisting of recursive ranker elimination of correlated window features, within-window wavelengths, and clustered window features chose the subset with the fewest features at six. With these six bands, QDA was able to obtain a mean training F1 score of about 0.86 and a test F1 score of about 0.70. The pipeline identified an optimal set of six bands for nitrogen estimation in table grape leaves, centered at 596, 611, 401, 1619, 825, and 1386 nm, respectively (Fig. 4).

TABLE IV. FEATURE SELECTION METHOD COMPARISON FOR NITROGEN DETECTION

| Feature Selection Method | Subset size (features) | F1 mean (extreme N) | F1 std (extreme N) | F1 (moderate N) |
|---|---|---|---|---|
| None | 51 | 0.93 | 0.03 | 0.69 |
| SelectKBest | 6 | 0.67 | 0.07 | 0.57 |
| SVM-RFE | 10 | 0.88 | 0.02 | 0.74 |
| ReliefF | 4 | 0.64 | 0.05 | 0.62 |
| Random Forest | 4 | 0.68 | 0.06 | 0.60 |
| LASSO | 9 | 0.86 | 0.03 | 0.67 |
| CCSA | 9 | 0.85 | 0.03 | 0.77 |
| Full Ensemble | 7 | 0.86 | 0.05 | 0.68 |
| Recursive Ranker Elimination | 7 | 0.87 | 0.03 | 0.68 |
| Full Pipeline | 6 | 0.86 | 0.02 | 0.70 |

## IV. DISCUSSION

Correlation windowing was able to replace each group of highly correlated adjacent wavelengths with a single, averaged, representative feature in the first stage of the pipeline, greatly reducing the feature set size from 1,339 wavelengths to just 51 correlated windows before further processing. The full ensemble of six rankers dramatically reduced the feature set size from 51 to just 7 correlated window features while sacrificing only 0.07 from the mean training F1 score and only 0.01 from the test F1 score. The winning ensemble of 3 rankers from recursive ranker elimination, Random Forest, LASSO, and ReliefF, also selected 7 features, but improved the mean training F1 score by 0.01 while reducing standard deviation by 0.02 and not changing the test F1 score. The within-window selection and clustering stage of the pipeline successfully re-centered and standardized the widths of the 7 selected correlated windows from the first stage toward more optimal positions and practical bandwidths. The resulting clustered windows had bandwidths of about 10, 20, or 40 nm and overlapping clusters were fused so that the newly formed bands were not overlapping.

The final stage of the pipeline selected the optimal set of clustered window features, which resulted in a feature set size of just six spectral bands scoring a mean F1 training score of 0.86 (and accuracy score of 0.87) on the 720 extreme nitrogen samples and an improved F1 test score of 0.70 (and accuracy score of 0.72) on the 227 moderate nitrogen samples. For comparison, Cerovic et al. [31] obtained a classification accuracy score of 0.83 on a set of 59 wine grape leaves using the nitrogen balance index (NBI), but with a single threshold separating low and high classes at 40 mg/g N. Even though training and feature selection were strictly performed on extreme nitrogen samples, the classification performance using the selected features on the unseen moderate nitrogen test samples was quite good, which is a promising sign for the informativeness of the selected bands for nitrogen detection more generally. The benefits of converting high resolution hyperspectral data to multispectral sensor bands through feature selection include (i) reduced model complexity, (ii) reduced risk of overfitting, (iii) reduced runtime, computation, and storage needed, (iv) increased robustness, and (v) reduced sensor cost [6].

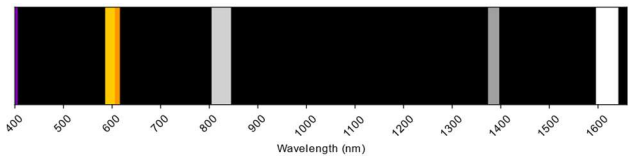

Fig. 4. Optimal set of spectral bands for nitrogen detection selected by the full pipeline.

Some potential sources of error in the study included the limited sensor range, relatively small sample size, nonuniform leaf sampling from each vine, and sharing of the ground truth label for nitrogen among all leaves from the same vine. One of the selected bands ranged from 400 to 406 nm with a center of 401 nm, meaning it was restricted at 400 nm on the lower end due to the lower cutoff of the sensor used. The training set used for this study contained one hyperspectral point measurement from each leaf sample, whereas hyperspectral and multispectral images can contain thousands of measurements in the form of pixels. The leaves for this study were not collected from the same nodes of each vine, but from single shoots collected from each vine. The ground truth labels of nitrogen found through chemical analysis were evaluated on mixed samples of multiple leaves from the same vine, rather than on individual leaves. This label sharing could have overlooked the differences in leaf nitrogen throughout the leaves on the same vine. Future work will involve applying the pipeline on a large aerial dataset of hyperspectral imagery to validate the selected bands on a larger sample size at the canopy scale.

## V. Conclusion

This paper presents a pipeline with several stages of ensemble feature selection and wavelength clustering which was applied to a hyperspectral dataset of leaves from 'Flame Seedless' grapevines in California under different fertilization regimes to produce an optimal set of bands for classifying leaf nitrogen level with a QDA classifier. The pipeline selected six bands centered at 596, 611, 401, 1619, 825, and 1386 nm with bandwidths of 20, 10, 6, 46, 40, and 23 nm, respectively, from the 1,339 original bands. The selected yellow-orange visible bands centered at 596 and 611 nm lie in a blind spot between the green and red bands of typical commercial multispectral sensors. Likewise, the selected violet visible band centered at 401 nm and the selected shortwave infrared bands centered at 1619 and 1386 nm lie outside the range of commercial sensors. Designing a custom multispectral sensor with the selected bands may offer a performance improvement over the standard UAV mountable sensors for remote sensing of table grape nitrogen. The pipeline can also aid in the design of application-specific multispectral sensors for detecting different target variables and for different application domains.

## Acknowledgment

The authors would like to thank Dr. Matthew W. Fidelibus of the Department of Viticulture and Enology and German Zuniga-Ramirez of the Department of Biological and Agricultural Engineering at the University of California, Davis for making available the hyperspectral data and ground truth labels. The authors would also like to thank the California Table Grape Commission for financially supporting the sampling and hyperspectral data collection costs.